\documentclass[twocolumn]{article}
\usepackage[utf8]{inputenc}

\title{Speech Commands: A Dataset for Limited-Vocabulary Speech Recognition }
\author{Pete Warden\\
Google Brain\\
Mountain View, California\\
{\tt\small petewarden@google.com}
}
\date{April 2018}

\usepackage[square,sort,comma,numbers]{natbib}
\usepackage{graphicx}
\usepackage{listings}
\usepackage{hyperref} 
\usepackage[T1]{fontenc}

\begin{document}

\maketitle

\section{Abstract}

Describes an audio dataset\citep{speechcommandsv2} of spoken words designed to help train and evaluate keyword spotting systems. Discusses why this task is an interesting challenge, and why it requires a specialized dataset that’s different from conventional datasets used for automatic speech recognition of full sentences. Suggests a methodology for reproducible and comparable accuracy metrics for this task. Describes how the data was collected and verified, what it contains, previous versions\citep{speechcommandsv1} and properties. Concludes by reporting baseline results of models trained on this dataset.

\section{Introduction}
Speech recognition research has traditionally required the resources of large organizations such as universities or corporations to pursue. People working in those organizations usually have free access to either academic datasets through agreements with groups like the Linguistic Data Consortium\citep{ldcwebsite}, or to proprietary commercial data.

As speech technology has matured, the number of people who want to train and evaluate recognition models has grown beyond these traditional groups, but the availability of datasets hasn’t widened. As the example of ImageNet\citep{imagenet_cvpr09} and similar collections in computer vision has shown, broadening access to datasets encourages collaborations across groups and enables apples-for-apples comparisons between different approaches, helping the whole field move forward.

The Speech Commands dataset is an attempt to build a standard training and evaluation dataset for a class of simple speech recognition tasks. Its primary goal is to provide a way to build and test small models that detect when a single word is spoken, from a set of ten or fewer target words, with as few false positives as possible from background noise or unrelated speech. This task is often known as keyword spotting.

To reach a wider audience of researchers and developers, this dataset has been released under the Creative Commons BY 4.0 license\citep{cc4}. This enables it to easily be incorporated in tutorials and other scripts where it can be downloaded and used without any user intervention required (for example to register on a website or email an administrator for permission). This license is also well known in commercial settings, and so can usually be dealt with quickly by legal teams where approval is required.

\section{Related Work}
Mozilla's Common Voice dataset\citep{commonvoice} has over 500 hours from 20,000 different people, and is available under the Creative Commons Zero license (similar to public domain). This licensing makes it very easy to build on top of. It is aligned by sentence, and was created by volunteers reading requested phrases through a web application.

LibriSpeech\citep{vassil2015librispeech} is a collection of 1,000 hours of read English speech, released under a Creative Commons BY 4.0 license, and stored using the open source FLAC encoder, which is widely supported. Its labels are aligned at the sentence level only, thus lacking word-level alignment information. This makes it more suitable for full automatic speech recognition than keyword spotting.

TIDIGITS\citep{tidigits} contains 25,000 digit sequences spoken by 300 different speakers, recorded in a quiet room by paid contributors. The dataset is only available under a commercial license from the Language Data Consortium, and is stored in the NIST SPHERE file format, which proved hard to decode using modern software. Our initial experiments on keyword spotting were performed using this dataset.

CHiME-5\citep{chime5} has 50 hours of speech recorded in people’s homes, stored as 16 KHz WAV files, and available under a restricted license. It’s aligned at the sentence level.

\section{Motivations}

Many voice interfaces rely on keyword spotting to start interactions. For example you might say "Hey Google" or "Hey Siri"\citep{heysiri} to begin a query or command for your phone. Once the device knows that you want to interact, it’s possible to send the audio to a web service to run a model that’s only limited by commercial considerations, since it can run on a server whose resources are controlled by the cloud provider. The initial detection of the start of an interaction is impractical to run as a cloud-based service though, since it would require sending audio data over the web from all devices all the time. This would be very costly to maintain, and would increase the privacy risks of the technology.

Instead, most voice interfaces run a recognition module locally on the phone or other device. This listens continuously to audio input from microphones, and rather than sending the data over the internet to a server, they run models that listen for the desired trigger phrases. Once a likely trigger is heard, the transfer of the audio to a web service begins. Because the local model is running on hardware that’s not under the web service provider’s control, there are hard resource constraints that the on-device model has to respect. The most obvious of these is that the mobile processors typically present have total compute capabilities that are much lower than most servers, so to run in near real-time for an interactive response, on-device models must require fewer calculations than their cloud equivalents. More subtly, mobile devices have limited battery lives and anything that is running continuously needs to be very energy efficient or users will find their device is drained too quickly. This consideration doesn’t apply to plugged-in home devices, but those do have thermal constraints on how much heat they can dissipate that restrict the amount of energy available to local models, and are encouraged by programs like EnergyStar to reduce their overall power usage as much as possible. A final consideration is that users expect a fast response from their devices, and network latency can be highly variable depending on the environment, so some initial acknowledgement that a command was received is important for a good experience, even if the full server response is delayed.

These constraints mean that the task of keyword spotting is quite different to the kind of speech recognition that’s performed on a server once an interaction has been spotted:

\begin{itemize}
\item Keyword spotting models must be smaller and involved less compute.
\item They need to run in a very energy-efficient way.
\item Most of their input will be silence or background noise, not speech, so false positives on those must be minimized.
\item Most of the input that is speech will be unrelated to the voice interface, so the model should be unlikely to trigger on arbitrary speech.
\item The important unit of recognition is a single word or short phrase, not an entire sentence.
\end{itemize}

These differences mean that the training and evaluation process between on-device keyword spotting and general speech recognition models is quite different. There are some promising datasets to support general speech tasks, such as Mozilla’s Common Voice, but they aren’t easily adaptable to keyword spotting.

This Speech Commands dataset aims to meet the special needs around building and testing on-device models, to enable model authors to demonstrate the accuracy of their architectures using metrics that are comparable to other models, and give a simple way for teams to reproduce baseline models by training on identical data. The hope is that this will speed up progress and collaboration, and improve the overall quality of models that are available.

A second important audience is hardware manufacturers. By using a publicly-available task that closely reflects product requirements, chip vendors can demonstrate the accuracy and energy usage of their offerings in a way that’s easily comparable for potential purchasers. This increased transparency should result in hardware that better meets product requirements over time. The models should also provide clear specifications that hardware engineers can use to optimize their chips, and potentially suggest model changes that make it easier to provide efficient implementations. This kind of co-design between machine learning and hardware can be a virtuous circle, increasing the flow of useful information between the domains in a way that helps both sides.

\section{Collection}

\subsection{Requirements}

I made the decision to focus on capturing audio that reflected the on-device trigger phrase task described above. This meant that the use of studio-captured samples seemed unrealistic, since that audio would lack background noise, would be captured with high-quality microphones, and in a formal setting. Successful models would need to cope with noisy environments, poor quality recording equipment, and people talking in a natural, chatty way. To reflect this, all utterances were captured through phone or laptop microphones, wherever users happened to be. The one exception was that I asked them to avoid recording themselves whenever there were background conversations happening for privacy reasons, so I asked them to be in a room alone with the door closed.

I also decided to focus on English. This was for pragmatic reasons, to limit the scope of the gathering process and make it easier for native speakers to perform quality control on the gathered data. I hope that transfer learning and other techniques will still make this dataset useful for other languages though, and I open-sourced the collection application to allow others to easily gather similar data in other languages. I did want to gather as wide a variety of accents as possible however, since we’re familiar from experience with the bias towards American English in many voice interfaces.

Another goal was to record as many different people as I could. Keyword-spotting models are much more useful if they’re speaker-independent, since the process of personalizing  a model to an individual requires an intrusive user interface experience. With this in mind, the recording process had to be quick and easy to use, to reduce the number of people who would fail to complete it.

I also wanted to avoid recording any personally-identifiable information from contributors, since any such data requires handling with extreme care for privacy reasons. This meant that I wouldn’t ask for any attributes like gender or ethnicity, wouldn’t require a sign-in through a user ID that could link to personal data, and would need users to agree to a data-usage agreement before contributing.

To simplify the training and evaluation process, I decided to restrict all utterances to a standard duration of one second. This excludes longer words, but the usual targets for keyword recognition are short so this didn’t seem to be too restrictive. I also decided to record only single words spoken in isolation, rather than as part of a sentence, since this more closely resembles the trigger word task we’re targeting. It also makes labeling much easier, since alignment is not as crucial.

\subsection{Word Choice}

I wanted to have a limited vocabulary to make sure the capture process was lightweight, but still have enough variety for models trained on the data to potentially be useful for some applications. I also wanted the dataset to be usable in comparable ways to common proprietary collections like TIDIGITS. This led me to pick twenty common words as the core of our vocabulary. These included the digits zero to nine, and in version one, ten words that would be useful as commands in IoT or robotics applications; "Yes", "No", "Up", "Down", "Left", "Right", "On", "Off", "Stop", and "Go". In version 2 of the dataset, I added four more command words; “Backward”, “Forward”, “Follow”, and “Learn”.
One of the most challenging problems for keyword recognition is ignoring speech that doesn’t contain triggers, so I also needed a set of words that could act as tests of that ability in the dataset. Some of these, such as “Tree”, were picked because they sound similar to target words and would be good tests of a model’s discernment. Others were chosen arbitrarily as short words that covered a lot of different phonemes. The final list was "Bed", "Bird", "Cat", "Dog", "Happy", "House", "Marvin", "Sheila", "Tree", and "Wow".

\subsection{Implementation}

To meet all these requirements, I created an open-source web-based application that recorded utterances using the WebAudioAPI\citep{webaudioapi}. This API is supported on desktop browsers like Firefox and Chrome, and on Android mobile devices. It’s not available on iOS, which was considered to be unfortunate but there were no alternatives that were more attractive. I also looked into building native mobile applications for iOS and Android, but I found that users were reluctant to install them, for privacy and security reasons. The web experience requires users to grant permission to the website to access the microphone, but that seemed a lot more acceptable, based on the increased response rate. The initial test of the application was hosted at an appspot.com subdomain, but it was pointed out that teaching users to give microphone permissions to domains that were easy for malicious actors to create was a bad idea. To address this, the final home of the application was moved to: \begin{lstlisting}[frame=tlrb]
https://aiyprojects.withgoogle.com/open_speech_recording
\end{lstlisting}
This is a known domain that’s controlled by Google, and so it should be much harder to create confusing spoofs of.

The initial page that a new user sees when navigating to the application explains what the project is doing, and asks them to explicitly and formally agree to participating in the study. This process was designed to ensure that the resulting utterances could be freely redistributed as part of an open dataset, and that users had a clear understanding of what the application was doing. When a user clicks on “I Agree”, a session cookie is added to record their agreement. The recording portion of the application will only be shown if this session cookie is found, and all upload accesses are guarded by cross-site request forgery tokens, to ensure that only audio recorded from the application can be uploaded, and that utterances are from users who have agreed to the terms.

The recording page asks users to press a “Record” button when they’re ready, and then displays a random word from the list described above. The word is displayed for 1.5 seconds while audio is recorded, and then another randomly-chosen word is shown after a one-second pause. Each audio clip is added to a list that’s stored locally on the client’s machine, and they remain there until the user has finished recording all words and has a chance to review them. The random ordering of words was chosen to avoid pronunciation changes that might be caused by repetition of the same word multiple times. Core words are shown five times each in total, whereas auxiliary words only appear once. There are 135 utterances collected overall, which takes around six minutes in total to run through completely. The user can pause and restart at any point.

Once the recording process is complete, the user is asked to review all of the clips, and if they’re happy with them, upload them. This then invokes a web API which uploads the audio to the server application, which saves them into a cloud storage bucket. The WebAudioAPI returns the audio data in OGG-compressed format, and this is what gets stored in the resulting files. The session ID is used as the prefix of each file name, and then the requested word is followed by a unique instance ID for the recording. This session ID has been randomly generated, and is not tied to an account or any other demographic information, since none has been generated. It does serve as a speaker identifier for utterances however. To ensure there’s a good distribution of different speakers, once a user has gone through this process once a cookie is added to the application that ensures they can’t access the recording page again.

To gather volunteers for this process, I used appeals on social media to share the link and the aims of the project. I also experimented with using paid crowdsourcing for some of the utterances, though the majority of the dataset comes from the open site.

\subsection{Quality Control}

The gathered audio utterances were of variable quality, and so I needed criteria to accept or reject submissions. The informal guideline I used was that if a human listener couldn’t tell what word was being spoken, or it sounded like an incorrect word, then the clip should be rejected. To accomplish this, I used several layers of review.

To remove clips that were extremely short or quiet, I took advantage of the nature of the OGG compression format. Compressed clips that contained very little audio would be very small in size, so a good heuristic was that any files that were smaller than 5 KB were unlikely to be correct. To implement this rule, I used the following Linux shell command:

\begin{lstlisting}[frame=tlrb, language=bash]
find ${BASEDIR}/oggs -iname "*.ogg" -size -5k -delete
\end{lstlisting}

With that complete, I then converted the OGG files into uncompressed WAV files containing PCM sample data at 16KHz, since this is any easier format for further processing:

\begin{lstlisting}[frame=tlrb, language=bash]
find ${BASEDIR}/oggs -iname "*.ogg" -print0 | xargs -0 basename -s .ogg | xargs -I {} ffmpeg -i ${BASEDIR}/oggs/{}.ogg -ar 16000 ${BASEDIR}/wavs/{}.wav
\end{lstlisting}

Samples from other sources came as varying sample-rate WAV files, so they were also resampled to 16 KHz WAV files using a similar ffmpeg command.

\subsection{Extract Loudest Section}

From manual inspection of the results, there were still large numbers of utterances that were too quiet or completely silent. The alignment of the spoken words within the 1.5 second file was quite arbitrary too, depending on the speed of the user’s response to the word displayed. To solve both these problems, I created a simple audio processing tool called Extract Loudest Section to examine the overall volume of the clips. As a first stage, I summed the absolute differences of all the samples from zero (using a scale where -32768 in the 16-bit sample data was -1.0 as a floating-point number, and +32767 was 1.0), and looked at the mean average of that value to estimate the overall volume of the utterance. From experimentation, anything below 0.004 on this metric was likely to be to quiet to be intelligible, and so all of those clips were removed.

To approximate the correct alignment, the tool then extracted the one-second clip that contained the highest overall volume. This tended to center the spoken word in the middle of the trimmed clip, assuming that the utterance was the loudest part of the recording. To run these processes, the following commands were called:

\begin{lstlisting}[frame=tlrb, language=bash]
git clone https://github.com/petewarden/extract_loudest_section tmp/extract_loudest_section
cd tmp/extract_loudest_section
make
cd ../..
mkdir -p ${BASEDIR}/trimmed_wavs
/tmp/extract_loudest_section/gen/bin/extract_loudest_section ${BASEDIR}'/wavs/*.wav' ${BASEDIR}/trimmed_wavs/
\end{lstlisting}

\subsection{Manual Review}
These automatic processes caught technical problems with quiet or silent recordings, but there were still some utterances that were of incorrect words or were unintelligible for other reasons. To filter these out I turned to commercial crowdsourcing. The task asked workers to type in the word they heard from each clip, and gave a list of the expected words as examples. Each clip was only evaluated by a single worker, and any clips that had responses that didn’t match their expected labels were removed from the dataset.

\subsection{Release Process}

The recorded utterances were moved into folders, with one for each word. The original 16-digit hexadecimal speaker ID numbers from the web application’s file names were hashed into 8-digit hexadecimal IDs. Speaker IDs from other sources (like the paid crowdsourcing sites) were also hashed into the same format. This was to ensure that any connection to worker IDs or other personally-identifiable information was removed. The hash function used is stable though, so in future releases the IDs for existing files should remain the same, even as more speakers are added.

\subsection{Background Noise}
A key requirement for keyword spotting in real products is distinguishing between audio that contains speech, and clips that contain none. To help train and test this capability, I added several minute-long 16 KHz WAV files of various kinds of background noise. Several of these were recorded directly from noisy environments, for example near running water or machinery. Others were generated mathematically using these commands in Python:

\begin{lstlisting}[frame=tlrb, language=Python]
scipy.io.wavfile.write('/tmp/white_noise.wav', 16000, np.array(((acoustics.generator.noise(16000*60, color='white'))/3) * 32767).astype(np.int16))
scipy.io.wavfile.write('/tmp/pink_noise.wav', 16000, np.array(((acoustics.generator.noise(16000*60, color='pink'))/3) * 32767).astype(np.int16))
\end{lstlisting}

To distinguish these files from word utterances, they were placed in a specially-named \verb|"_background_noise_"| folder, in the root of the archive.

\section{Properties}

The final dataset consisted of 105,829 utterances of 35 words, broken into the categories and frequencies shown in Table 1.

\begin{figure}[h!]
\begin{center}
\begin{tabular}{ | c | c | }
\hline
Word & Number of Utterances \\
\hline
Backward & 1,664 \\
Bed & 2,014 \\
Bird & 2,064 \\
Cat & 2,031 \\
Dog & 2,128 \\
Down & 3,917 \\
Eight & 3,787 \\
Five & 4,052 \\
Follow & 1,579 \\
Forward & 1,557 \\
Four & 3,728 \\
Go & 3,880 \\
Happy & 2,054 \\
House & 2,113 \\
Learn & 1,575 \\
Left & 3,801 \\
Marvin & 2,100 \\
Nine & 3,934 \\
No & 3,941 \\
Off & 3,745 \\
On & 3,845 \\
One & 3,890 \\
Right & 3,778 \\
Seven & 3,998 \\
Sheila & 2,022 \\
Six & 3,860 \\
Stop & 3,872 \\
Three & 3,727 \\
Tree & 1,759 \\
Two & 3,880 \\
Up & 3,723 \\
Visual & 1,592 \\
Wow & 2,123 \\
Yes & 4,044 \\
Zero & 4,052 \\
\hline
\end{tabular}
\caption{How many recordings of each word are present in the dataset}
\label{Table 1}
\end{center}
\end{figure}

Each utterance is stored as a one-second (or less) WAVE format file, with the sample data encoded as linear 16-bit single-channel PCM values, at a 16 KHz rate. There are 2,618 speakers recorded, each with a unique eight-digit hexadecimal identifier assigned as described above. The uncompressed files take up approximately 3.8 GB on disk, and can be stored as a 2.7GB gzip-compressed tar archive.

\section{Evaluation}

One of this dataset’s primary goals is to enable meaningful comparisons between different models’ results, so it’s important to suggest some precise testing protocols. As a starting point, it’s useful to specify exactly which utterances can be used for training, and which must be reserved for testing, to avoid overfitting. The dataset download includes a text file called \verb|validation_list.txt|, which contains a list of files that are expected to be used for validating results during training, and so can be used frequently to help adjust hyperparameters and make other model changes. The \verb|testing_list.txt| file contains the names of audio clips that should only be used for measuring the results of trained models, not for training or validation. The set that a file belongs to is chosen using a hash function on its name. This is to ensure that files remain in the same set across releases, even as the total number changes, so avoid set cross-contamination when trying old models on the more recent test data. The Python implementation of the set assignment algorithm is given in the TensorFlow tutorial code\citep{whichset} that is a companion to the dataset.

\subsection{Top-One Error}

The simplest metric to judge a trained model against is how many utterances it can correctly identify. In principle this can be calculated by running the model against all the files in the testing set, and comparing the reported against the expected label for each. Unlike image classification tasks like ImageNet, it’s not obvious how to weight all of the different categories. For example, I want a model to indicate when no speech is present, and separately to indicate when it thinks a word has been spoken that’s not one it recognizes. These “open world” categories need to be weighted according to their expected occurrence in a real application to produce a realistic metric that reflects the perceived quality of the results in a product.

The standard chosen for the TensorFlow speech commands example code is to look for the ten words "Yes", "No", "Up", "Down", "Left", "Right", "On", "Off", "Stop", and "Go", and have one additional special label for “Unknown Word”, and another for “Silence” (no speech detected). The testing is then done by providing equal numbers of examples for each of the twelve categories, which means each class accounts for approximately 8.3\% of the total. The "Unknown Word" category contains words randomly sampled from classes that are part of the target set. The "Silence" category has one-second clips extracted randomly from the background noise audio files.

I've uploaded a standard set of test files\citep{speechcommandstestsetv2} to make it easier to reproduce this metric. If you want to calculate the canonical Top-One error for a model, run inference on each audio clip, and compare the top predicted class against the ground truth label encoded in its containing subfolder name. The proportion of correct predictions will give you the Top-One error. There's also a similar collection of test files\citep{speechcommandstestsetv1} available for version one of the dataset.

The example training code that accompanies the dataset\citep{tutorial} provides results of 88.2\% on this metric for the highest-quality model when fully trained. This translates into a model that qualitatively gives a reasonable, but far from perfect response, so it’s expected that this will serve as a baseline to be exceeded by more sophisticated architectures.

\subsection{Streaming Error Metrics}

Top-One captures a single dimension of the perceived quality of the results, but doesn’t reveal much about other aspects of its performance in a real application. For example, models in products receive a continuous stream of audio data and don’t know when words start and end, whereas the inputs to Top One evaluations are aligned to the beginning of utterances. The equal weighting of each category in the overall score also doesn’t reflect the distribution of trigger words and silence in typical environments.

To measure some of these more complex properties of models, I test them against continuous streams of audio and score them on multiple metrics. Here's what the baseline model trained with V2 data produces:

\begin{lstlisting}[frame=tlrb, language=bash]
49.0% matched, 46.0% correctly, 3.0% wrongly, 0.0% false positives
\end{lstlisting}

To produce this result, I ran the following bash script against the 10 minute streaming test audio clip and ground truth labels:

\begin{lstlisting}[frame=tlrb, language=bash]
bazel run tensorflow/examples/speech_commands:freeze -- --start_checkpoint=/tmp/speech_commands_train/conv.ckpt-18000 --output_file=/tmp/v2_frozen_graph.pb
bazel run tensorflow/examples/speech_commands:test_streaming_accuracy -- --graph=/tmp/v2_frozen_graph.pb --wav=/tmp/speech_commands_train/streaming_test.wav --labels=/tmp/speech_commands_train/conv_labels.txt --ground_truth=/tmp/speech_commands_train/streaming_test_labels.txt
\end{lstlisting}

\begin{itemize}
\item Matched-percentage represents how many words were correctly identified, within a given time tolerance.
\item Wrong-percentage shows how many words were correctly distinguished as speech rather than background noise, but were given the wrong class label.
\item False-positive percentage is the number of words detected that were in parts of the audio where no speech was actually present.
\end{itemize}

An algorithm for calculating these values given an audio file and a text file listing ground truth labels is implemented in TensorFlow as \verb|test_streaming_accuracy.cc|\citep{teststreamingaccuracy}.

Performing successfully on these metrics requires more than basic template recognition of audio clips. There has to be at least a very crude set of rules to suppress repeated recognitions of the same word in short time frames, so default logic for this is implemented in \verb|recognize_commands.cc|\citep{recognizecommands}.

This allows a simple template-style recognition model to be used directly to generate these statistics. One of the other configurable features of the accuracy test is the time tolerance for how close to the ground truth’s time a recognition result must be to count as a match. The default for this is set to 750ms, since that seems to match with requirements for some of the applications that are supported.

To make reproducing and comparing results easier, I've made available a one-hour audio file\citep{speechcommandsstreamingtestv2} containing a mix of utterances at random times and noise, together with a text file marking the times and ground truth labels of each utterance. This was generated using the script included in the TensorFlow tutorial, and can be used to compare different models performance on streaming applications.

\subsection{Historical Evaluations}

Version 1 of the dataset\citep{speechcommandsv1} was released August 3rd 2017, and contained 64,727 utterances from 1,881 speakers. Training the default convolution model from the TensorFlow tutorial (based on Convolutional Neural Networks for Small-footprint Keyword Spotting\citep{cnnkws}) using the V1 training data gave a Top-One score of 85.4\%, when evaluated against the test set from V1.
Training the same model against version 2 of the dataset\citep{speechcommandsv2}, documented in this paper, produces a model that scores 88.2\% Top-One on the training set extracted from the V2 data. A model trained on V2 data, but evaluated against the V1 test set gives 89.7\% Top-One, which indicates that the V2 training data is responsible for a substantial improvement in accuracy over V1. The full set of results are shown in Table 2.

\begin{figure}[h!]
\begin{center}
\begin{tabular}{ | c | c | c | }
\hline
Data & V1 Training & V2 Training \\
\hline
V1 Test & 85.4\% & 89.7\% \\
\hline
V2 Test & 82.7\% & 88.2\% \\
\hline
\end{tabular}
\caption{Top-One accuracy evaluations using different training data}
\label{Table 2}
\end{center}
\end{figure}

These figures were produced using the checkpoints produced by the following training commands:
\begin{lstlisting}[frame=tlrb, language=bash]
python tensorflow/examples/speech_commands/train.py --data_url=http://download.tensorflow.org/data/speech_commands_v0.01.tar.gz
python tensorflow/examples/speech_commands/train.py --data_url=http://download.tensorflow.org/data/speech_commands_v0.02.tar.gz
\end{lstlisting}

The results of these commands are available as pretrained checkpoints\citep{speechcommandscheckpoints}. The evaluations were performed by running variations on the following command line (with the v1/v2's substituted as appropriate):

\begin{lstlisting}[frame=tlrb, language=bash]
python tensorflow/examples/speech_commands/train.py --data_url=http://download.tensorflow.org/data/speech_commands_v0.0{1,1}.tar.gz --start_checkpoint=${HOME}/speech_commands_checkpoints/conv-v{1,2}.ckpt-18000
\end{lstlisting}

\subsection{Applications}

The TensorFlow tutorial gives a variety of baseline models, but one of the goals of the dataset is to enable the creation and comparison of a wide range of models on a lot of different platforms, and version one of has enabled some interesting applications. CMSIS-NN\citep{cmsisnn} covers a new optimized implementation of neural network operations for ARM microcontrollers, and uses Speech Commands to train and evaluate the results. Listening to the World\citep{listeningtotheworld} demonstrates how combining the dataset and UrbanSounds\citep{urbansounds} can improve the noise tolerance of recognition models. Did you Hear That\citep{didyouhearthat} uses the dataset to test adversarial attacks on voice interfaces. Deep Residual Learning for Small Footprint Keyword Spotting\citep{deepresidual} shows how approaches learned from ResNet can produce more efficient and accurate models. Raw Waveform-based Audio Classification\citep{rawwaveform} investigates alternatives to traditional feature extraction for speech and music models. Keyword Spotting Through Image Recognition\citep{keywordspotting} looks at the effect virtual adversarial training on the keyword task.

\section{Conclusion}

The Speech Commands dataset has shown to be useful for training and evaluating a variety of models, and the second version shows improved results on equivalent test data, compared to the original.

\section{Acknowledgements}

Massive thanks are due to everyone who donated recordings to this data set, I'm very grateful. I also couldn't have put this together without the help and support of Billy Rutledge, Rajat Monga, Raziel Alvarez, Brad Krueger, Barbara Petit, Gursheesh Kour, Robert Munro, Kirsten Gokay, David Klein, Lukas Biewald, and all the AIY and TensorFlow teams.

\onecolumn{
\bibliographystyle{IEEEtran}
\bibliography{references}
}
\end{document}